\documentclass[a4paper, 10pt, conference]{IEEEtran}
\IEEEoverridecommandlockouts
\usepackage{cite}
\usepackage{amsmath,amssymb,amsfonts}
\usepackage{algorithmic}
\usepackage{graphicx}
\usepackage{textcomp}
\usepackage{lipsum}
\usepackage{multirow}
\usepackage{xcolor}
\def\BibTeX{{\rm B\kern-.05em{\sc i\kern-.025em b}\kern-.08em
    T\kern-.1667em\lower.7ex\hbox{E}\kern-.125emX}}
\begin{document}

\title{Agentic AI for Intent-Based Industrial Automation\\
\thanks{This work was partially supported by Coordenação de Aperfeiçoamento de Pessoal de Nível Superior - Brasil (CAPES) -- Financial Code 001, by Fundação de Amparo à Pesquisa do Estado de São Paulo (FAPESP) - grant \#2020/09838-0, and the Conselho Nacional de Desenvolvimento Científico e Tecnológico (CNPq) - grant \#311380/2021-2.}
}

\author{\IEEEauthorblockN{Marcos Lima Romero}
\IEEEauthorblockA{\textit{Center for Engineering, Modeling and} \\
\textit{Applied Social Sciences}\\ 
\textit{Federal University of ABC - UFABC}\\
Santo André, Brazil \\
https://orcid.org/0009-0008-1406-6349}
\and

\IEEEauthorblockN{Ricardo Suyama}
\IEEEauthorblockA{\textit{Center for Engineering, Modeling and}\\
\textit{Applied Social Sciences}\\
\textit{Federal University of ABC - UFABC}\\
Santo André, Brazil \\
https://orcid.org/0000-0002-8398-5268}
}

\maketitle

\begin{abstract}
The recent development of Agentic AI systems, empowered by autonomous large language models (LLMs) agents with planning and tool-usage capabilities, enables new possibilities for the evolution of industrial automation and reduces the complexity introduced by Industry 4.0. This work proposes a conceptual framework that integrates Agentic AI with the intent-based paradigm, originally developed in network research, to simplify human–machine interaction (HMI) and better align automation systems with the human-centric, sustainable, and resilient principles of Industry 5.0.
Based on the intent-based processing, the framework allows human operators to express high-level business or operational goals in natural language, which are decomposed into actionable components. These intents are broken into expectations, conditions, targets, context, and information that guide sub-agents equipped with specialized tools to execute domain-specific tasks. A proof of concept was implemented using the CMAPSS dataset and Google Agent Developer Kit (ADK), demonstrating the feasibility of intent decomposition, agent orchestration, and autonomous decision-making in predictive maintenance scenarios. The results confirm the potential of this approach to reduce technical barriers and enable scalable, intent-driven automation, despite data quality and explainability concerns. 
\end{abstract}

\begin{IEEEkeywords}
agentic ai, intent-based, industry 5.0, manufacturing, automation
\end{IEEEkeywords}

\section{Introduction}
Since the advent of Industry 4.0 in the mid-2010s, the volume of data generated by industrial enterprises has grown exponentially. A recent study by ABI Research projects that annual data production in the industrial sector will increase from 1.9 zettabytes (ZB) to 4.4 ZB by 2030 \cite{abi}. Cyber-physical systems have bridged the physical and digital worlds, enabling machines, products, and management systems to communicate and exchange data in real time through Industrial Internet of Things (IIoT) networks. As a result, production processes have become highly automated, optimized through Big Data analytics and AI, and capable of dynamically adapting to changes. Mass customization and on-demand production have become increasingly feasible, while data-driven predictive maintenance strategies help prevent unexpected downtime \cite{passalacqua2025human}.

However, although these technological advances have significantly improved industrial capabilities, they have also introduced new layers of complexity for human operators, who must now manage massive volumes of data and oversee the operation of increasingly sophisticated systems \cite{operator4}. Industry 5.0 introduces a shift toward a human-centric approach to address this emerging gap.

Based on the ongoing adoption of the principles of Industry 4.0, Industry 5.0, as defined by the European Commission, proposes new essential pillars for the next industrial era: human-centric approach, sustainability, and resilience \cite{european}. It emphasizes the adoption of human-centered strategies, wherein technology is not designed to replace human labor but to augment and sustainably enhance human capabilities. This vision is closely aligned with the United Nations Sustainable Development Goals (SDGs), particularly Goals 8, 9, and 12 — Decent Work and Economic Growth; Industry, Innovation, and Infrastructure; and Responsible Consumption and Production \cite{hassan2024systematic}.

In this context, AI emerges as a key enabler of Industry 5.0, helping to bridge the gap between the increasing complexity of systems and the need for more intuitive human interactions. Through natural language communication, Large Language Models (LLMs) have gained widespread adoption, allowing users to perform complex tasks with just a few rounds of prompt-based conversation \cite{figlie2024towards}. Building on this advancement, LLM-based agents are now being developed with enhanced capabilities, including reasoning, planning, and the ability to use specialized tools to autonomously execute specific tasks \cite{lim2024large}. These tasks can range from simple system status inquiries to the orchestration of complex business or operational intent requests.

In line with the human-centric vision of Industry 5.0, communication based on intentions becomes a fundamental paradigm. Rather than requiring users to provide step-by-step instructions focusing on \textit{how} to do, intent-based communication enables them to express \textit{what} they want to achieve clearly and naturally. The system interprets these high-level intents and autonomously determines the optimal actions to fulfill them \cite{zeydan2024generative}. This paradigm shift aims to make interactions with complex technological environments more intuitive, efficient, and resilient, reducing operational errors and supporting continuous adaptation to evolving goals and operational scenarios.

Despite these advances, traditional human-machine interaction (HMI) within industrial environments still presents significant challenges. Interfaces are often complex, not user-friendly, and require extensive training for operators to avoid costly errors. As previously discussed, the increasing complexity of industrial systems tends to be mirrored in their HMIs, which continue to accumulate new functions, data streams, and control mechanisms. Consequently, operators must attain ever-higher levels of specialization to interact effectively with these systems. Without support from more intuitive interaction models and an AI-driven abstraction layer, managing vast amounts of data and monitoring multiple subsystems simultaneously becomes nearly unfeasible.

This article aims to present LLM agents operating under an intent-based paradigm, radically simplifying HMIs and streamlining engagement with industrial processes. By shifting the focus from command-driven interfaces to intent-based communication, these agents are capable of interpreting user goals expressed in natural language and autonomously orchestrating the necessary actions to fulfill them. The main contributions of this article can be summarized as:
\begin{itemize}
    \item A conceptual framework that leverages LLM-based agents to enable intent-driven interaction with industrial automation systems, in alignment with the principles of Industry 5.0.

    \item A novel intention-processing pipeline that translates natural language inputs into structured, actionable industrial tasks using expectations, conditions, targets, context, and information.

    \item A practical proof of concept, based on the CMAPSS dataset \cite{saxena2008damage} and Google Agent Developer Kit (ADK), demonstrating the feasibility of agentic orchestration in a predictive maintenance scenario.
\end{itemize}

The article is organized as follows: section \ref{background} provides an overview of AI agents and Agentic AI, traditional human-machine interaction challenges, and intent-based systems. Building on this foundation, section \ref{framework} outlines a novel architecture that places an LLM agent at the core of an intention-processing pipeline, supported by custom tools designed to interface with industrial data and systems. In the section \ref{concept}, a realistic industrial scenario is presented using the CMAPSS dataset to demonstrate the feasibility and effectiveness of the proposed approach. This is followed by the section \ref{discussion} with a discussion of the implications and benefits of intent-based agents, as well as challenges and future directions for research and development. Finally, section \ref{conclusion} summarizes the contributions and highlights the potential of LLM-based intent systems within the context of Industry 5.0.

\section{Background}\label{background}

With the advancement of data-driven and automated solutions urged by Industry 4.0, particularly through machine learning (ML) and advanced data analytics, AI naturally emerges as a means to improve decision-making and enable more efficient automation, considering the human-centric approach advocated by Industry 5.0. However, the use of AI in the industrial and manufacturing context is not a novelty.

The application of AI in industrial automation has been discussed in the literature since the 1970s \cite{nitzan1976programmable, minsky1976automation}, initially through simpler concepts such as perceptrons. The use of AI agents in factories has been proposed since the early 1990s \cite{chandrasekaran1990aaai}, where researchers recognized that the true value lay in the collaboration of a network of agents, including AI agents, human agents, machine systems, and sensors.

\subsection{AI Agents vs. Agentic AI}

An AI agent is a software element capable of acting autonomously for a user or system to execute tasks \cite{bawcom2024agentic}. Since the early 1990s, AI agents have been recognized as key components of intelligent automation systems, with foundational research highlighting the importance of distributed collaboration among human operators, machine systems, and software agents \cite{chandrasekaran1990aaai}. Initial implementations predominantly relied on rule-based systems and symbolic reasoning, which, while groundbreaking at the time, were limited in their adaptability to dynamic industrial environments. 

These early AI agents, driven by the lack of computational power at the time, typically operated under strict guidelines: they required explicit instructions, handled short-term or narrowly defined tasks, responded only to direct commands, and could only be updated through manual reprogramming. As such, they were most effective in predictable, well-structured settings. 

Over the subsequent decades, advances in ML, neural networks, multi-agent systems, and reinforcement learning progressively enhanced the autonomy, adaptability, and scalability of AI agents \cite{tzafestas1995artificial}. Building on these developments, the advent of LLMs has enabled a new workflow known as \textit{Agentic AI}, where agents are empowered with capabilities such as natural language understanding, reasoning, planning, and collaboration with other agents \cite{ferrag2025llm}. Table \ref{tab:agent_vs_agentic} shows a comparison between AI Agents and Agentic AI. 

\begin{table*}[hbt]
\centering
\caption{Comparison Between AI Agents and Agentic AI}
\label{tab:agent_vs_agentic}
\begin{tabular}{|p{4cm}|p{5cm}|p{7cm}|}
\hline
\textbf{Aspect} & \textbf{AI Agent} & \textbf{Agentic AI} \\
\hline
Autonomy & Operates under strict human-defined rules & Acts independently with minimal human input \\
\hline
Instruction & Requires specific commands and step-by-step instructions & Understands and interprets high-level intents \\
\hline
Task Scope & Focused on short-term, well-defined tasks & Oriented toward long-term, dynamic, and complex goals \\
\hline
Adaptability & Limited; updates require reprogramming & Continuously learns and adapts from experience \\
\hline
Decision Making & Based on predefined logic or rules & Capable of reasoning, planning, and multi-factor decision-making \\
\hline
Environment Handling & Performs best in predictable environments & Designed to operate in uncertain and evolving environments \\
\hline
Interaction Style & Reactive to direct inputs & Proactive, capable of initiating actions and managing workflows \\
\hline
Tool Use & Limited or static tool integration & Can autonomously select and use tools or delegate tasks to sub-agents \\
\hline
Example Technologies & Rule-based systems, symbolic AI, early expert systems & LLM-based agents, multi-agent coordination, intent-based workflows \\
\hline
\end{tabular}
\end{table*}

\begin{figure*}[hbt]
    \centering
    \includegraphics[width=1\linewidth]{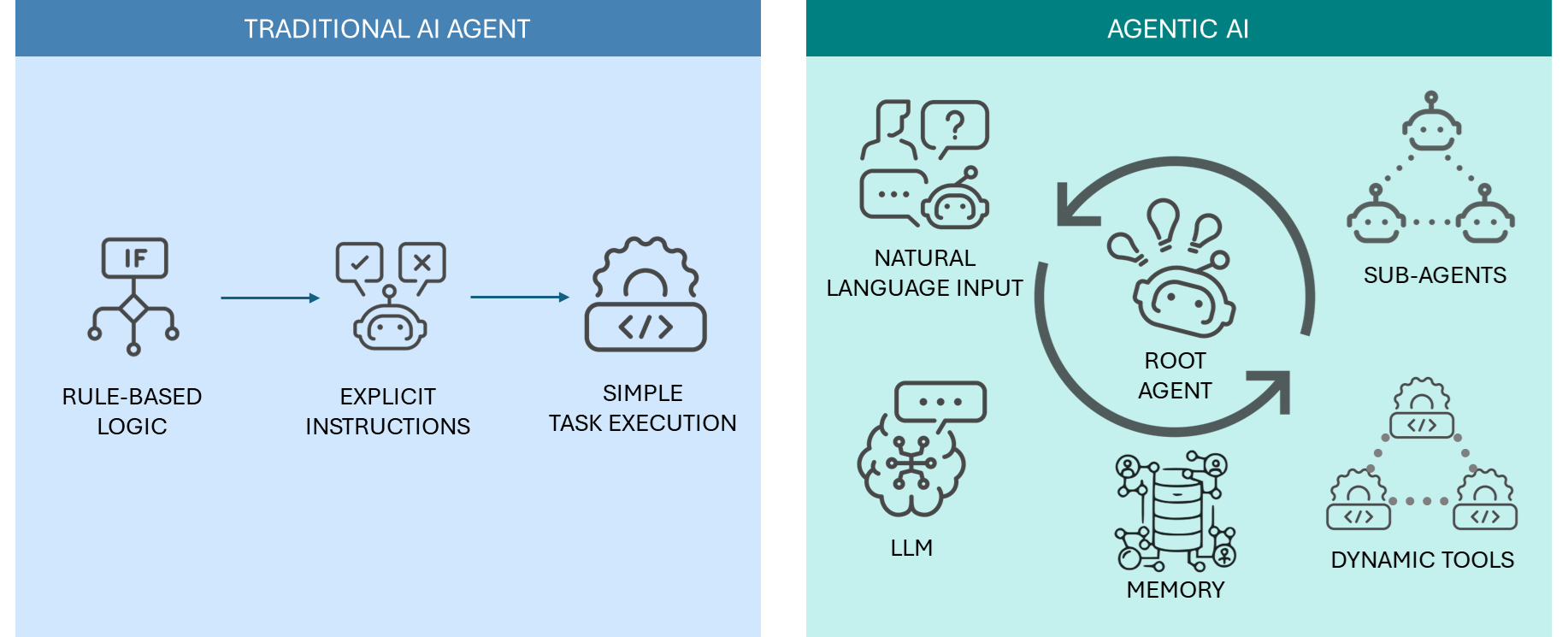}
    \caption{Traditional AI Agent vs. Agentic AI}
    \label{fig:agentic}
\end{figure*}

Unlike traditional agents, Agentic AI systems operate with minimal human intervention, pursue long-term or intent-based goals through adaptive strategies, continuously learn from experience, and make context-aware decisions, with local or external memory, that consider multiple factors simultaneously \cite{hughes2025ai}, as shown in Fig. \ref{fig:agentic}. These agents can interpret the semantics of natural language input and delegate tasks to sub-agents, whether Small Language Models (SLMs), other LLMs, or domain-specific tools, while operating within data-restricted environments to ensure privacy and security. In doing so, Agentic AI bridges the gap between generative AI and action-oriented execution, offering a flexible and autonomous solution for complex and evolving industrial applications \cite{acharya2025agentic}.

Some Agentic AI design patterns are emerging as blueprints for constructing intelligent agents in complex environments \cite{rashmiranjan2025empirical, bousetouane2025agentic, e2025rag, bornet2025agentic}. The most influential are:
\begin{itemize}
    \item[] \textbf{ReAct} (Reasoning and Acting), which integrates step-by-step reasoning with LLMs and real-time action execution with tools.
    \item[] \textbf{CodeAct} builds on this by enabling agents to generate and execute code on the fly, making them capable of solving technical problems or interacting with systems.
    \item[] \textbf{Modern Tool Use} structures to choose and utilize specialized tools with standard communications such as Model Context Protocol (MCP) or Agent to Agent (A2A), including external third-party tools.
    \item[] \textbf{Self-Reflection} introduces cognitive loops, allowing agents to critique and revise their past actions via learning to improve future performance.
    \item[] \textbf{Multi-Agent} workflows orchestrate collaborative or competitive interactions between multiple agents with specialized roles, promoting modularity, scalability, and emergent behavior in complex problem-solving.
    \item[] \textbf{Agentic RAG} (Retrieval-Augmented Generation) uses external knowledge bases dynamically through agents that search, evaluate, and synthesize information.
\end{itemize}
Selecting the appropriate design pattern depends on several factors, including available computational resources, task complexity, and critical considerations such as privacy and security.

\subsection{Human-Machine Interaction}

Human–Machine Interaction (HMI) focuses on developing user-friendly and efficient interfaces that enable seamless and intuitive communication between humans and machines. Achieving this requires a deep understanding of both human behavior and the functional limitations of machines. 

Over time, HMI has evolved alongside technological advancements. Early interfaces relied on punched cards or command-line inputs, demanding users learn specific machine protocols. The advent of graphical user interfaces (GUIs) made interactions more intuitive and visually accessible. Subsequent innovations, such as touchscreens and voice recognition, further personalized the experience by allowing more natural modes of input. Despite this progress, a key challenge remains: ensuring that HMIs are easy to use, transparent, and accessible to a wide range of users \cite{mourtzis2023future}. To address this, the integration of natural language processing via LLMs has emerged as a promising enhancement, allowing users to express what they want to achieve through intentions without needing deep technical knowledge of how to implement them.

Recent studies have explored the integration of LLMs into industrial automation systems, enhancing HMIs and enabling more adaptive control mechanisms \cite{lim2024large, figlie2024towards, gantayat2025efficiency, keskin2025llm, brown2025human}. These efforts align with the principles of Industry 5.0 and highlight the high potential of LLMs in managing complex industrial tasks. However, most of these works fall short of implementing intent-driven architectures, and the full potential of agentic workflows remains largely unexploited. Although, they provide a foundational layer upon which intent-based systems and Agentic AI can be further developed for industrial automation.

\subsection{Intent Based Systems}

Thanks to advances in computational power and the emergence of LLMs, a new technological concept has been enabled: intent-based systems. Leveraging the natural language capabilities of these models, human users can interact with systems by expressing what they intend to achieve, rather than detailing step-by-step instructions on how to achieve it. This paradigm allows human operators to focus on higher-level reasoning tasks, closer to the business layer, while delegating execution to an agentic workflow, abstracting the technical layer, focusing on a parametric agnostic conversation with minimal external intervention \cite{acharya2025agentic}.

The concept of intent-based systems was initially explored in the context of telecommunications networks, \textit{intent} representing an evolved version of \textit{policy} in the network \cite{zeydan2020recent}. Telecommunications is one of the most complex fields to operate in, due to the diversity of applications, tools, and vendors involved. Ensuring seamless connectivity and mobility for millions of users, without noticeable interruptions or failures, is a significant challenge. To address this, a growing body of research has recently focused on the development of intent-based networking solutions.

Since 2016, intent-based networks have emerged as a promising approach to automate and self-orchestrate the complex systems of telecommunications infrastructure \cite{schulz2016intent}. The concept has since gained significant traction, with a search on Google Scholar yielding over 400 publications containing the exact phrase ``intent-based networks.'' Within the network domain, this paradigm has been explored in diverse verticals, including vehicular networks \cite{shen2024intent, safavat2020elliptic}, healthcare systems \cite{njah2023toward}, and industrial automation \cite{tomur2023intent, ustok2022asset}.

Recent research efforts have explored the application of intent-based paradigms in industrial automation. One such approach proposed an intent-based management framework aimed at enabling end-to-end automation across industrial systems \cite{baktir2024intent}. The authors extended the common intent model defined by TM Forum \cite{tm-forum}, originally designed for autonomous networks, to address the specific requirements of industrial environments. Another notable initiative introduced the use of fine-tuned LLMs to bring intent-based interaction to the shop floor, focusing on adapting these models to accurately interpret and act on industrial intents expressed in natural language \cite{zeydan2024generative}.

Although the concept of intent-based interactions in industrial applications is relatively recent, current implementations still lack enabling advanced workflows that fully leverage the potential of an Agentic AI architecture. The key differentiator of this perspective lies in its ability to translate high-level intentions into concrete planning and actionable tasks with LLM-based agents. 


\section{Proposed Framework}\label{framework}

\begin{figure*}[hbt]
    \centering
    \includegraphics[width=1\linewidth]{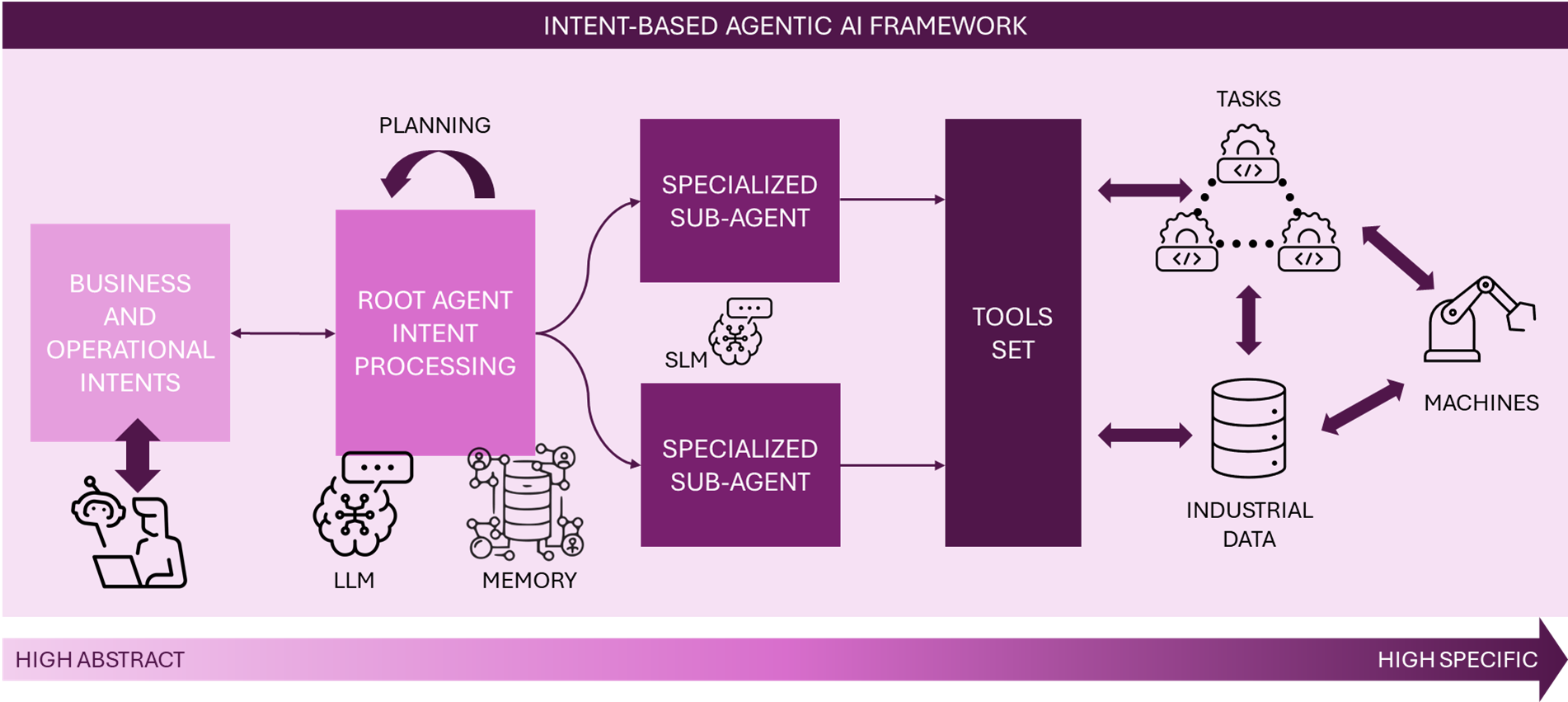}
    \caption{Proposed framework for Industry 5.0 applying intent-based and Agentic AI.}
    \label{fig:framework}
\end{figure*}

Bringing together the concepts of Agentic AI, the evolution of HMI interactions, the human-centric focus of Industry 5.0, and intent-based processing, this work aims to contribute to the existing literature by advancing the state-of-the-art in industrial automation. The proposed framework is presented in Fig. \ref{fig:framework}.

\subsection{Architecture}
The selected design pattern for the architecture is the \textbf{multi-agent model}, which enables agent cooperation while maintaining a hierarchy between a root agent, responsible for processing user intent, and sub-agents that handle interactions with the specific domains of the industrial application.

The architecture follows this workflow: initially, the user provides input in natural language, expressing business or operational intentions by focusing on the desired outcomes rather than the specific technical execution. In doing so, there is an abstraction of the intent layer from the execution layer.

Once the intent is defined, the root agent processes it using an LLM, along with stored memory from previous interactions or knowledge bases such as knowledge graphs. This allows the agent to reason about the optimal plan of action and delegate tasks accordingly. The root agent is in charge of generating the action plan, including defining the steps and iterations necessary to achieve the intended outcome.

Following delegation by the root agent, specialized sub-agents can utilize LLMs or SLMs fine-tuned for industrial contexts to determine next steps, such as interacting with other agents or invoking tools to perform specific tasks, e.g., real-time data collection, system configuration, or sending commands to machines.

The set of tools provides agents with a library of possible actions to interact with the specific industrial environment. These tools are modular and can be developed according to each machine or system protocol. As a result, this set is dynamic and adaptable to the needs of a given operation or business goal. Adding a new function is straightforward due to the modular structure; once developed, the new tool can simply be made available for agent use.

\subsection{Intention Processing}
Intention processing plays a central role in the proposed framework. Based on the interpretation of the user's natural language input, the LLM must reason and decompose the intention into several components, as shown in Table \ref{tab:intention_components}: \textbf{expectations}, \textbf{conditions}, \textbf{targets}, \textbf{resources}, \textbf{context}, and \textbf{information} \cite{tm-forum}.

\begin{table}[hbt]
\centering
\caption{Key Components of Intention Processing}
\label{tab:intention_components}
\centering
\resizebox{\linewidth}{!}{%
\begin{tabular}{|c|p{5cm}|}
\hline
\textbf{Component} & \textbf{Description} \\
\hline
\textbf{Expectations} & Define what is required or expected from the system. Core elements of an intent may relate to performance, behavior, or service delivery. \\
\hline
\textbf{Conditions} & Logical expressions used to evaluate whether an expectation is being met. Typically based on measurable criteria like performance indicators or system states. They determine the compliance status of expectations. \\
\hline
\textbf{Targets} & Specify the \textbf{resources} or entities to which the intent applies. Can be defined statically (explicit list) or dynamically (using filters or criteria). \\
\hline
\textbf{Context} & Provides additional information such as priority, timeframes, or environmental scope. Helps interpret when and how expectations should be applied. \\
\hline
\textbf{Information} & Includes auxiliary data not directly useful for guiding decisions, such as customer IDs, related intents, or operational hints. \\
\hline
\end{tabular}%
}
\end{table}

With this decomposition, it is possible to make an execution plan to achieve the desired expectations to the appropriate targets, without risk of non-compliance according to conditions, context, and information. The root agent can decide whether to use specialized sub-agents to conclude its tasks or not; eventually, in some architectures, the root agent can call tools directly.

\subsection{Sub-agents}
Based on the plan decomposed by the root agent, sub-agents may be triggered to carry out specific tasks or retrieve the required information. These sub-agents can take various forms, including other LLM-based agents, smaller and more specialized SLM agents, or even non-LLM-based agents tailored for domain-specific operations or system orchestration.

The structure of the agentic workflow can vary depending on the application context. Agent interactions may follow different patterns, such as iterative loops until predefined criteria are met, sequential delegation of tasks, or parallel execution to enhance performance and responsiveness.

With more specialized context and targeted instructions, sub-agents can address tasks more effectively to achieve the desired outcomes. They have access to a defined set of tools, and since each tool includes well-described input/output specifications, sub-agents can autonomously select the most appropriate tool for the task at hand.

\subsection{Custom Tools}
Tools are modular components that extend the agent's ability to interact programmatically with external systems, execute tasks, and retrieve or process information beyond its internal reasoning capabilities. These tools function as callable units, such as code functions, API connectors, or simulation interfaces, that the agent can dynamically select and invoke during task execution. Tools do not reason on their own; instead, the root agent's LLM determines which tool to use and provides the appropriate input.

The process by which agents use the tools is dynamic and structured. According to the agent's input, the appropriate tool is selected based on its description, invoked with generated arguments, observing the output, and incorporating the result into further decision making. This allows agents to link multiple tools together or repeat operations based on conditions, making them highly adaptable and effective for completing complex tasks using intent-driven workflows.

Different types of tools support varying operational needs, including custom-defined functions, built-in utilities, such as search or code execution, long-running asynchronous tools, and integrations with third-party libraries. To ensure effectiveness, tools must have clear function names and descriptions, and instructions should specify how the agent should respond to different outcomes, handling errors, or combining tools in sequence. In industrial environments, this allows agents to interact with real-time data sources or control machines, enabling actions that would not be possible through agents based on language models alone.

\section{Proof of Concept}\label{concept}

A proof of concept (PoC) is proposed to validate the aforementioned framework. Using the well-known industrial dataset CMAPSS and Python's open-source libraries, the objective is to develop a reproducible blueprint for applying Agentic AI in industrial environments. 

The prototype focuses on integrating intent-based interaction, an LLM-based root agent for reasoning and planning, and specialized sub-agents capable of orchestrating domain-specific tasks such as diagnostics, data querying, and maintenance planning. This PoC aims not only to demonstrate technical feasibility but also to provide insights into architectural patterns, workflow orchestration, and tool integration strategies for agent-based industrial systems. Ultimately, the project lays the groundwork for future research and development of intelligent, adaptive, and intent-driven automation aligned with the principles of Industry 5.0.

\subsection{Dataset}
The CMAPSS \cite{saxena2008damage} is an industrial, well-known, synthetic dataset used to train and test predictions of remaining useful life (RUL). It was created by a simulation tool of the same name (Commercial Modular Aero-Propulsion System Simulation) coded in MATLAB\textsuperscript{\textregistered} and Simulink\textsuperscript{\textregistered}. 

Each line of the dataset contains an instance of an engine represented by a set of measurements, including three operational settings and 21 sensor readings collected at each cycle. The primary prediction objective is to estimate the RUL, the number of cycles remaining before the engine reaches failure. Table \ref{tab:cmapss_attributes} shows a list of parameters and their units \cite{asif2022deep}. 

\begin{table}[h]
\centering
\caption{CMAPSS Dataset Attributes}
\label{tab:cmapss_attributes}
\begin{tabular}{|c|c|c|}
\hline
\textbf{Attribute} & \textbf{Unit} & \textbf{Type} \\
\hline
Engine ID & - & \multirow{2}{*}{Index} \\
Cycle & - & \\
\hline
Speed & Ma & \multirow{3}{*}{Operational Setting} \\
Altitude & feet & \\
Sea level temperature & °F & \\
\hline
Fan inlet temperature & °R & \multirow{21}{*}{Sensor}  \\
LPC outlet temperature & °R & \\
HPC outlet temperature & °R & \\
LPT outlet temperature & °R & \\
Fan inlet pressure & psia & \\
Bypass-duct pressure & psia & \\
HPC outlet pressure & psia & \\
Physical fan speed & rpm & \\
Physical core speed & rpm & \\
Engine pressure ratio & - & \\
HPC outlet static pressure & psia & \\
Ratio of fuel flow & pps/psia & \\
Corrected fan speed & rpm & \\
Corrected core speed & rpm & \\
Bypass ratio & - & \\
Burner fuel-air ratio & - & \\
Bleed enthalpy & - & \\
Required fan speed & rpm & \\
Required fan conversion speed & rpm & \\
High-pressure turbines cool air flow & lbm/s & \\
Low-pressure turbines cool air flow & lbm/s & \\
\hline
\end{tabular}
\end{table}

Widely used in the development and evaluation of Prognostics and Health Management (PHM) models for predictive maintenance, the dataset has become a benchmark in the field, with over 3,000 academic articles referencing it on Google Scholar. However, in the context of this work, the dataset is not used for predictive modeling. Instead, it serves as a structured and realistic source of industrial data to demonstrate the system's ability to interact with a dynamic environment and execute agentic workflows.

\subsection{Implementation}

Several libraries have gained traction in both industry and academia for the implementation of Agentic AI systems, including LangChain\footnote{https://www.langchain.com/}, CrewAI\footnote{https://www.crewai.com/}, and Smolagents\footnote{https://smolagents.org/}. For this project, the recently released Google ADK\footnote{https://google.github.io/adk-docs/} was selected. The choice was motivated by its flexible orchestration capabilities via AutoFlow, native integration with Gemini\footnote{https://ai.google.dev/gemini-api/docs} (free tier of Gemini 2.0 Flash), and a built‑in web interface that eliminates the need to develop a front‑end from scratch. Additionally, GitHub Codespaces\footnote{https://github.com/features/codespaces} was used as the execution environment, provisioning a cloud‑based virtual machine with 2 CPU cores, 8 GB of RAM, and 32 GB of storage.

\begin{figure}[hbt]
    \centering
    \includegraphics[width=1\linewidth]{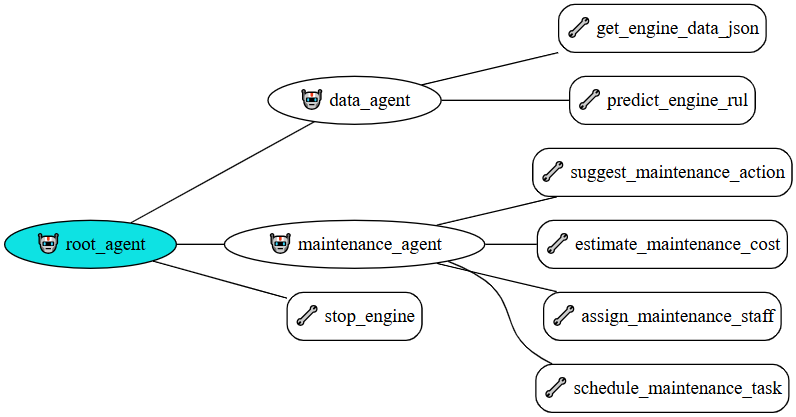}
    \caption{Agentic AI implementation with Google ADK.}
    \label{fig:implementation}
\end{figure}

To reduce complexity and avoid excessive agent calls, a controlled subset of the CMAPSS dataset was used, containing 20 engines. The code, available on GitHub\footnote{https://github.com/RomeroCode/talk-to-your-factory}, performs a multi-agent architecture as illustrated in Figure \ref{fig:implementation}. 

A \texttt{root\_agent} first decomposes the user’s intent into the core components: \emph{Expectations}, \emph{Conditions}, \emph{Targets}, \emph{Context} and \emph{Information}. After reasoning and planning, it can delegate tasks to two specialized sub‑agents defined for this PoC:

\begin{itemize}
  \item \texttt{data\_agent}: Retrieves engine telemetry and provides RUL predictions. In this PoC, RUL values are taken directly from the dataset (ground truth), since predictive modeling is outside of the current scope. It has access to the tools \texttt{get\_engine\_data\_json} and \texttt{predict\_engine\_rul}.
  \item \texttt{maintenance\_agent}: Identifies engines near failure and plans preventive maintenance: scheduling downtime, assigning teams, and estimating per‑engine workload. It has the following tools available to invoke:
  \begin{itemize}
      \item \texttt{suggest\_maintenance\_action} \item \texttt{estimate\_maintenance\_cost}  \item \texttt{assign\_maintenance\_staff}
      \item \texttt{schedule\_maintenance\_task}
  \end{itemize}
  
\end{itemize}

Upon receiving outputs from the \texttt{data\_agent}, the \texttt{root\_agent} autonomously decides whether to invoke the \texttt{maintenance\_agent} for preventive action or to initiate a shutdown of critical engines by invoking the \texttt{stop\_engine} tool directly. The \texttt{root\_agent} also returns the results as a structured table summary, including all stops and maintenance schedules.

\subsection{Results}

To validate the system’s ability to decompose high-level intentions into actionable steps, the following initial prompt was submitted to the agent, without any prior context: 

\begin{quote}
\textit{"I need to maintain all engines working well according to their predicted RUL, avoiding unexpected stops, please make a consolidated predictive maintenance plan in a table format."}
\end{quote}

Based on this input, asking for the \texttt{root\_agent} how it processed the intent and structured it, it answered that it decomposed into the following components:

\begin{itemize}
  \item \textbf{Expectation:} Maintain all engines in optimal operating condition to prevent unexpected failures. This was inferred from the phrase ``maintain all the engines working well according to their predicted RUL, avoiding unexpected stops,'' implying continuous, reliable engine performance.
  
  \item \textbf{Targets:} All 20 engines in the fleet. The request for a consolidated plan covering ``all the engines'' indicates that the intent applies to the entire fleet under management.
  
  \item \textbf{Conditions:} RUL must be above a critical threshold, as defined by the \texttt{maintenance\_agent}. This stems from the clause ``according to their predicted RUL,'' which implies that engines with RULs below the threshold should be prioritized for maintenance.
  
  \item \textbf{Context:} High-priority, proactive maintenance strategy aimed at minimizing unplanned downtime and maximizing engine lifespan. The phrasing of the request reflects a strategic and preventative approach.
  
  \item \textbf{Information:} Predicted RUL values and engine sensor data, obtained from the \texttt{data\_agent}, are required to evaluate each engine's condition and inform decision-making.
\end{itemize}

Based on this decomposition, the agentic system performed a coordinated reasoning process between agents, evaluated the statuses of the engine, and generated a predictive maintenance plan. The final output of the system, detailing the actions to be taken for each engine, is summarized in Table \ref{tab:engine_actions}.

\begin{table*}[hbt]
\caption{Summary of Engine Maintenance Actions}
\label{tab:engine_actions}
\centering
\resizebox{\linewidth}{!}{%
\renewcommand{\arraystretch}{1.2}
\begin{tabular}{|c|c|c|c|c|c|c|c|}
\hline
\textbf{\# Engines} & \textbf{RUL Range} & \textbf{Recommended Action} & \textbf{Priority} & \textbf{Cost (USD)} & \textbf{Labor Hours} & \textbf{Assigned Staff} & \textbf{Scheduled Time} \\
\hline
15 & 82--124 & MONITOR & low & 0 & 0 & [jr\_mechanic] & Within 7 days \\
\hline
1 & 69 & MONITOR & low & 0 & 0 & [jr\_mechanic] & Within 3 days \\
\hline
2 & 28, 50 & REPAIR & high & 6000 & 4 & [mechanic, jr\_mechanic] & Within 3 days \\
\hline
1 & 16 & STOP & critical & 15000 & 8 & [tech\_lead, sr\_mechanic] & IMMEDIATE \\
\hline
\end{tabular}%
}
\end{table*}

Additional tests were conducted in which multiple data points were queried, without requiring any operator's previous knowledge of engine protocols or operational manuals. The system responded correctly, demonstrating the usability and accessibility of an intent-based approach. This reinforces the potential of such systems to simplify human–machine interaction, allowing operators to focus on higher-level tasks that require business insight, strategic decision-making, or creative problem-solving.

\section{Discussion}\label{discussion}
Through natural language input, abstracting away technical complexity and aligning more closely with business needs, it was possible to successfully decompose an intent-based request into a concrete action plan to be delegated to specialized sub-agents. These sub-agents, equipped with modular tools adapted to the specific application domain, executed the tasks received from the \texttt{root\_agent} and returned outputs used to advance the planned workflow. This result fulfills the objective of simplifying the HMI and reinforces the human-centered approach promoted by Industry 5.0 and the United Nations SDGs, ensuring that the growing technical complexity of systems does not become a justification for replacing human labor with machines and AI.

The presented PoC and the underlying framework demonstrate the potential of Agentic AI in industrial automation. There remains a significant opportunity for further exploration, particularly through the integration of external sub-agents and tools to expand the range of actionable tasks in industrial contexts. The combination of the versatility of multi-agent systems with the conversational and associative power of LLMs points the way for innovative applications. These range from revisiting legacy concepts that were once constrained by limited technology to proposing entirely new and disruptive methods for various sectors of the industry.

Despite the enthusiasm, several challenges must be addressed before Agentic AI systems can be deployed at scale in production environments. These include concerns over information security, data privacy, energy expenditures, explainability of AI decisions, hallucination control, consistency and predictability of actions, effective prompt engineering, and, most critically, the quality of the underlying data. Overcoming these challenges will be essential to building trust and ensuring the reliability of Agentic AI in industrial applications. For example, in the context of security, some agentic architectures are already incorporating guardrail techniques to prevent sensitive information from leaking to external servers, helping enforce data boundaries and maintain compliance.

In future work, additional tests will be necessary to evolve the current PoC into a production-ready system. The use of real-world datasets and implementation in actual industrial environments will be essential to validate the framework and uncover any limitations. Alternative LLM architectures may also be explored, as well as fine-tuned SLMs specifically adapted for industrial automation contexts. As research and experimentation progress, benchmarking methodologies can be applied to evaluate system performance, accuracy, and the benefits brought by Agentic AI in real-world applications.

\section{Conclusion}\label{conclusion}
Although artificial intelligence has been a topic in industrial automation since the 1970s, and the use of AI agents has been reported since the 1990s, only now unprecedented computational power is available. Across all sectors, there is a growing trend to revisit legacy ideas once abandoned due to technical limitations. With the advent of Agentic AI, many of these ideas can now be approached with renewed potential and more promising results. Innovation alone does not drive adoption, but tangible benefits such as improved productivity, human dignity, and cost efficiency do.

This work demonstrates that it is possible to integrate AI in alignment with the three core pillars of Industry 5.0: a human-centric approach, sustainability, and resilience. By bringing industrial automation management and HMI closer to natural language, capable not only of understanding intent-based communication but also of planning actions using available tools, a new range of possibilities emerges. This enables human operators to focus on tasks requiring critical thinking and creativity, while delegating routine technical operations to an Agentic AI system. In doing so, humans concentrate on high-level objectives aligned with business goals, while the agentic system handles the technical layer.

Naturally, any emerging technology faces limitations. In the case of LLM-based agentic systems, prompt engineering becomes a critical factor. During testing, a certain sensitivity to prompt phrasing was observed, and small changes often led to different outcomes, highlighting the challenge of achieving predictability and consistency. In addition, the explainability problem is present in every AI application. Finally, data quality remains a concern, covering aspects such as reliability, accuracy, privacy, and trust. Addressing these challenges will be vital for large-scale deployment.

Future work can build upon this paper's important milestone in intent-based communication research through Agentic AI systems. With the continued momentum and enthusiasm around AI technologies, innovation is advancing rapidly and it is certain that new tools and techniques will soon emerge, bringing increasingly precise solutions to the advancement of industry.

\bibliographystyle{ieeetr}
\bibliography{main}

\begin{thebibliography}{10}

\bibitem{abi}
{ABI Research}, ``Industrial data generation forecast,'' Sep 2024.
\newblock [Online] Available on: https://www.abiresearch.com/news-resources/chart-data/manufacturing-industry-amount-of-data-generated.

\bibitem{passalacqua2025human}
M.~Passalacqua, R.~Pellerin, F.~Magnani, P.~Doyon-Poulin, L.~Del-Aguila, J.~Boasen, and P.-M. L{\'e}ger, ``Human-centred ai in industry 5.0: a systematic review,'' {\em International Journal of Production Research}, vol.~63, no.~7, pp.~2638--2669, 2025.

\bibitem{operator4}
D.~Romero, J.~Stahre, T.~Wuest, O.~Noran, P.~Bernus, A.~Fasth, Fast-Berglund, and D.~Gorecky, ``Towards an operator 4.0 typology: A human-centric perspective on the fourth industrial revolution technologies,'' 10 2016.

\bibitem{european}
E.~Commission, D.-G. for Research, Innovation, M.~Breque, L.~De~Nul, and A.~Petridis, {\em Industry 5.0 – Towards a sustainable, human-centric and resilient European industry}.
\newblock Publications Office of the European Union, 2021.

\bibitem{hassan2024systematic}
M.~A. Hassan, S.~Zardari, M.~U. Farooq, M.~M. Alansari, and S.~A. Nagro, ``Systematic analysis of risks in industry 5.0 architecture,'' {\em Applied Sciences}, vol.~14, no.~4, p.~1466, 2024.

\bibitem{figlie2024towards}
R.~Figli{\`e}, T.~Turchi, G.~Baldi, and D.~Mazzei, ``Towards an llm-based intelligent assistant for industry 5.0,'' in {\em Proceedings of the 1st International Workshop on Designing and Building Hybrid Human--AI Systems (SYNERGY 2024)}, vol.~3701, 2024.

\bibitem{lim2024large}
J.~Lim, B.~Vogel-Heuser, and I.~Kovalenko, ``Large language model-enabled multi-agent manufacturing systems,'' in {\em 2024 IEEE 20th International Conference on Automation Science and Engineering (CASE)}, pp.~3940--3946, IEEE, 2024.

\bibitem{zeydan2024generative}
E.~Zeydan, J.~Mangues, S.~S. Arslan, Y.~Turk, and M.~Liyanage, ``Generative artificial intelligence for intent-based industrial automation,'' {\em IEEE Consumer Electronics Magazine}, 2024.

\bibitem{saxena2008damage}
A.~Saxena, K.~Goebel, D.~Simon, and N.~Eklund, ``Damage propagation modeling for aircraft engine run-to-failure simulation,'' in {\em 2008 international conference on prognostics and health management}, pp.~1--9, IEEE, 2008.

\bibitem{nitzan1976programmable}
Nitzan, ``Programmable industrial automation,'' {\em IEEE Transactions on Computers}, vol.~100, no.~12, pp.~1259--1270, 1976.

\bibitem{minsky1976automation}
M.~Minsky, ``Automation and artificial intelligence,'' {\em Science, Technology, and the Modern Navy: Thirtieth Anniversary, 1946-1976}, vol.~37, p.~111, 1976.

\bibitem{chandrasekaran1990aaai}
B.~Chandrasekaran and F.~Coyle, ``Aaai'90,'' {\em IEEE Intelligent Systems}, vol.~5, no.~05, pp.~10--12, 1990.

\bibitem{bawcom2024agentic}
A.~Bawcom, N.~von Bismarck, A.~Tavakoli, H.~Harreis, C.~Giovine, J.~Kaplan, K.~Rowshankish, J.~Amar, L.~Yee, M.~Chui, R.~Roberts, L.~Hämäläinen, L.~Lerner, R.~Thomas, and V.~Chung, ``What is an ai agent?,'' tech. rep., McKinsey \& Company, Mar 2025.
\newblock [Online] Available on: https://www.mckinsey.com/featured-insights/mckinsey-explainers/what-is-an-ai-agent/.

\bibitem{tzafestas1995artificial}
S.~Tzafestas and H.~Verbruggen, {\em Artificial intelligence in industrial decision making, control, and automation: an introduction}.
\newblock Springer, 1995.

\bibitem{ferrag2025llm}
M.~A. Ferrag, N.~Tihanyi, and M.~Debbah, ``From llm reasoning to autonomous ai agents: A comprehensive review,'' {\em arXiv preprint arXiv:2504.19678}, 2025.

\bibitem{hughes2025ai}
L.~Hughes, Y.~K. Dwivedi, T.~Malik, M.~Shawosh, M.~A. Albashrawi, I.~Jeon, V.~Dutot, M.~Appanderanda, T.~Crick, R.~De’, {\em et~al.}, ``Ai agents and agentic systems: A multi-expert analysis,'' {\em Journal of Computer Information Systems}, pp.~1--29, 2025.

\bibitem{acharya2025agentic}
D.~B. Acharya, K.~Kuppan, and B.~Divya, ``Agentic ai: Autonomous intelligence for complex goals--a comprehensive survey,'' {\em IEEE Access}, 2025.

\bibitem{rashmiranjan2025empirical}
P.~D. Rashmiranjan, ``Empirical analysis of agentic ai design patterns in real-world applications,'' {\em International Journal of Innovative Research in Computer and Communication Engineering}, vol.~13, no.~4, pp.~9761--9769, 2025.

\bibitem{bousetouane2025agentic}
F.~Bousetouane, ``Agentic systems: A guide to transforming industries with vertical ai agents,'' {\em arXiv preprint arXiv:2501.00881}, 2025.

\bibitem{e2025rag}
G.~d.~A. e~Aquino, N.~d.~S. de~Azevedo, L.~Y.~S. Okimoto, L.~Y.~S. Camelo, H.~L. de~Souza~Bragan{\c{c}}a, R.~Fernandes, A.~Printes, F.~Cardoso, R.~Gomes, and I.~G. Torn{\'e}, ``From rag to multi-agent systems: A survey of modern approaches in llm development,'' 2025.

\bibitem{bornet2025agentic}
P.~Bornet, J.~Wirtz, T.~H. Davenport, D.~De~Cremer, B.~Evergreen, P.~Fersht, R.~Gohel, S.~Khiyara, P.~Sund, and N.~Mullakara, {\em Agentic Artificial Intelligence: Harnessing AI Agents to Reinvent Business, Work and Life}.
\newblock Irreplaceable Publishing, 2025.

\bibitem{mourtzis2023future}
D.~Mourtzis, J.~Angelopoulos, and N.~Panopoulos, ``The future of the human--machine interface (hmi) in society 5.0,'' {\em Future Internet}, vol.~15, no.~5, p.~162, 2023.

\bibitem{gantayat2025efficiency}
P.~K. Gantayat, T.~Kaur, M.~Majhi, U.~K. Jena, S.~Das, {\em et~al.}, ``From efficiency to innovation: Llm 5.0 and the future of industry,'' in {\em 2025 International Conference on Multi-Agent Systems for Collaborative Intelligence (ICMSCI)}, pp.~551--558, IEEE, 2025.

\bibitem{keskin2025llm}
Z.~Keskin, D.~Joosten, N.~Klasen, M.~Huber, C.~Liu, B.~Drescher, and R.~H. Schmitt, ``Llm-enhanced human-machine interaction for adaptive decision making in dynamic manufacturing process environments,'' {\em IEEE access}, 2025.

\bibitem{brown2025human}
B.~R. Brown, ``Human--machine teaming using large language models,'' in {\em Interdependent Human-Machine Teams}, pp.~41--66, Elsevier, 2025.

\bibitem{zeydan2020recent}
E.~Zeydan and Y.~Turk, ``Recent advances in intent-based networking: A survey,'' in {\em 2020 IEEE 91st Vehicular Technology Conference (VTC2020-Spring)}, pp.~1--5, IEEE, 2020.

\bibitem{schulz2016intent}
D.~Schulz, ``Intent-based automation networks: Toward a common reference model for the self-orchestration of industrial intranets,'' in {\em IECON 2016-42nd Annual Conference of the IEEE Industrial Electronics Society}, pp.~4657--4664, IEEE, 2016.

\bibitem{shen2024intent}
Y.~Shen, Y.~Ahn, M.~Gu, and J.~P. Jeong, ``Intent-based management for software-defined vehicles in intelligent transportation systems,'' in {\em 2024 IEEE 10th International Conference on Network Softwarization (NetSoft)}, pp.~1--6, IEEE, 2024.

\bibitem{safavat2020elliptic}
S.~Safavat and D.~B. Rawat, ``On the elliptic curve cryptography for privacy-aware secure aco-aodv routing in intent-based internet of vehicles for smart cities,'' {\em IEEE Transactions on Intelligent Transportation Systems}, vol.~22, no.~8, pp.~5050--5059, 2020.

\bibitem{njah2023toward}
Y.~Njah, A.~Leivadeas, J.~Violos, and M.~Falkner, ``Toward intent-based network automation for smart environments: A healthcare 4.0 use case,'' {\em IEEE Access}, vol.~11, pp.~136565--136576, 2023.

\bibitem{tomur2023intent}
E.~Tomur, Z.~Bilgin, U.~G{\"u}len, E.~U. Soykan, L.~Kara{\c{c}}ay, and F.~Karako{\c{c}}, ``Intent-based security for functional safety in cyber-physical systems,'' {\em IEEE Transactions on Emerging Topics in Computing}, 2023.

\bibitem{ustok2022asset}
R.~F. Ustok, A.~C. Baktir, and E.~D. Biyar, ``Asset administration shell as an enabler of intent-based networks for industry 4.0 automation,'' in {\em 2022 IEEE 27th International Conference on Emerging Technologies and Factory Automation (ETFA)}, pp.~1--8, IEEE, 2022.

\bibitem{baktir2024intent}
A.~C. Baktir and E.~D. Biyar, ``Intent-based management for industrial automation,'' in {\em 2024 IEEE 29th International Conference on Emerging Technologies and Factory Automation (ETFA)}, pp.~1--4, IEEE, 2024.

\bibitem{tm-forum}
{Autonomous Networks Project}, ``Intent common model,'' Technical Report (TR) 290, TM Forum, Jul. 2024.
\newblock Version 3.6.0. [Online] Available on: https://www.tmforum.org/resources/introductory-guide/intent-common-model-v3-6-0-tr290/.

\bibitem{asif2022deep}
O.~Asif, S.~A. Haider, S.~R. Naqvi, J.~F. Zaki, K.-S. Kwak, and S.~R. Islam, ``A deep learning model for remaining useful life prediction of aircraft turbofan engine on c-mapss dataset,'' {\em Ieee Access}, vol.~10, pp.~95425--95440, 2022.

\end{thebibliography}

\end{document}